# Low-cost foil/paper based touch mode pressure sensing element as artificial skin module for prosthetic hand

Rishabh B. Mishra[1, 2], Sherjeel M. Khan[1], Sohail F. Shaikh[1], Aftab M. Hussain[2], Muhammad M. Hussain[1, 3, *]

*Abstract*—Capacitive pressure sensors have several advantages in areas such as robotics, automation, aerospace, biomedical and consumer electronics. We present mathematical modelling, finite element analysis (FEA), fabrication and experimental characterization of ultra-low cost and paper-based, touch-mode, flexible capacitive pressure sensor element using Do-It-Yourself (DIY) technology. The pressure sensing element is utilized to design large-area electronics skin for low-cost prosthetic hands. The presented sensor is characterized in normal, transition, touch and saturation modes. The sensor has higher sensitivity and linearity in touch mode operation from 10 to 40 kPa of applied pressure compared to the normal (0 to 8 kPa), transition (8 to 10 kPa) and saturation mode (after 40 kPa) with response time of 15.85 ms. Advantages of the presented sensor are higher sensitivity, linear response, less diaphragm area, less von Mises stress at the clamped edges region, low temperature drift, robust structure and less separation gap for large pressure measurement compared to normal mode capacitive pressure sensors. The linear range of pressure change is utilized for controlling the position of a servo motor for precise movement in robotic arm using wireless communication, which can be utilized for designing skin-like structure for low-cost prosthetic hands.

## I. INTRODUCTION

This Pressure sensors are most commonly used in high performance devices for robotics [1-2], aerospace [3-4], industrial [5-6], biomedical [7-10], environmental monitoring [11-12], consumer and portable electronics applications [13-17]. The micro-electro-mechanical systems (MEMS)-based capacitive pressure sensor offers several advantages over piezo-resistive pressure sensor such as low temperature sensitivity, low power consumption, less side stress leading to long term stability, high pressure sensitivity and repeatability in output [4, 10, 18-20]. However, it requires large diaphragm area, has low dynamic range, is affected by parasitic capacitance and has non-linear response [4, 21]. Hence, the touch-mode capacitive pressure sensor was introduced to overcome these disadvantages. In normal mode capacitive pressure sensor, as shown in Fig 1(a), the diaphragm deflection must be kept 1/3rd of separation gap to avoid the pull-in phenomena. In this case, the diaphragm deflection follows the Kirchhoff's small deflection plate theory [4, 19-20]. However, in touch mode capacitive pressure sensor, the diaphragm touches the fixed bottom plate with a separation of the thin dielectric material [5, 18-21], as shown in Fig 1(b). In this mode, the diaphragm touch causes a major change in capacitance, while diaphragm deflection causes limited change. Thus, the touch mode capacitive sensor requires less separation gap between electrodes and less diaphragm area by which most of the disadvantages of MEMS-based capacitive pressure sensors can be eliminated [21-23]. There has been significant research on modelling, device optimization, sensitivity enhancement and achieving near-linear region of touch mode capacitive pressure sensor for various applications.

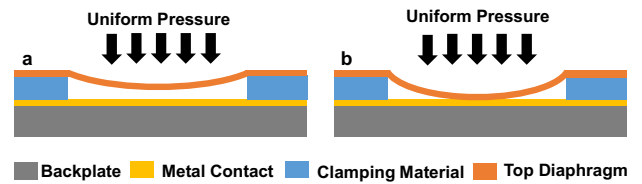

Fig. 1. Schematic of (a) Normal mode capacitive pressure sensor. (b) Touch mode capacitve pressure sensor.

The touch mode capacitive pressure sensor operates in four different modes: normal, transition, touch and saturation. The touch mode regime will have the maximum range of operation and will provide linear response. As pressure increases during touch mode operation, the capacitance varies linearly with pressure application and sensitivity of sensor will be much higher than in normal mode operation. In the saturation mode, bottom electrode of sensor provides support to the diaphragm for excessive high-pressure measurement, so that diaphragm does not break or crack [21-22].

With the advent of flexible electronics, the need for flexible capacitive pressure sensor came into existence for several applications like wearable electronics [16, 24], soft robotics [24-25], automobile [26], aerospace [27], and consumer electronics [28]. For designing a flexible pressure sensor, paper is a low-cost, light weight and easily available material [11, 24, 29]. It has good physical and chemical properties in terms of flexibility, foldability, and printability [11, 29-30]. The fabrication steps to design the paper sensors need very few and simple steps like photolithography, cutting, printing, etching, laser and plasma treatments [11, 29, 30-32].

Wang et al. 3D-printed an actuator using a layer of conductive polylactide (PLA) on a piece of paper for multiple applications in robotics, touch sensing, coloring and thermal imaging [25]. X. Liu *et al.* presented piezo-resistive pressure sensor after patterning silver and graphite inks as conductive material on the paper substrate and signal processing circuitry integrated on the same substrate, using which paper weighing balance was presented after connecting the piezo-resistor in Wheatstone's bridge fashion [29]. J. M. Nassar *et al.* presented pressure, temperature, pH and humidity sensors fabricated using aluminum foil/tapes, silver inks, sticky-notes, microfiber wipes/napkins, sponge, single/double side and double-sided tapes [11]. While designing capacitive pressure sensors using these materials, we have the options of microfiber wipe, sponge and air for dielectric materials. However, among these, electrodes separated by air dielectric have maximum sensitivity [11]. The capacitive sensing principle was implemented to fabricate paper-based capacitive touch pads using double-sided tape and conductive paper for book covers and beverages by A. D. Mazzeo *et al.* [31]. The capacitive sensing technique was utilized to design a wearable system for monitoring wheezing for asthma detection by S. Khan *et al.* using aluminum coated polyimide sheet and double-sided tape in which the edges of the circular, square and rectangular shape sensors are clamped by multi-layering of double-sided type [8]. The circular shape capacitive pressure sensor has maximum deflection among all the different shapes like elliptical, square, rectangular and pentagon shape, however the response of circular shape pressure sensor shows the highest non-linearity [32]. H. Oh et al. presented paper-based electronic devices for sensing, actuation, display and communication using crafts based digital sculpting and designed cube lamp, flowerpot, architectural model, smart toy with RFID tag and loudspeaker [33]. The pressure sensors for electronic skin can be utilized in the field of interactive electronics for implanting medical devices, artificial intelligence and designing robotic systems for several biomedical applications like prosthetic skins/hands, wearable devices for health monitoring [34-39]. The previous progress which has been in this direction consists to fabricate the touch mode capacitive pressure sensor requires cleanroom in which requires complex and sophisticated fabrication processes which is time consuming and requires expensive materials due to which the cost of sensor will be high [11, 29].

[1]R. B. Mishra, S. M. Khan, S. F. Shaikh and M. M. Hussain are associated with MMH Labs, Electrical Engineering, King Abdullah University of Science and Technology (KAUST), Thuwal 23955-6900, Kingdom of Saudi Arabia. [*email: muhammad.hussain@kaust.edu.sa)
[2]R. B. Mishra and A. M. Hussain are associated with Center for VLSI and Embedded Systems Technology (CVEST), International Institute of Information Technology (IIIT), Hyderabad, Telangana, 500032, India.
[3]M. M. Hussain is associated with Electrical Engineering and Computer Sciences (EECS), University of California, Berkeley, California 94720, USA. [email: mmhussain@berkely.edu].

In this paper, we present a low cost, flexible, aluminum-coated, polyimide paper-based, touch mode capacitive pressure sensor using DIY-electronics. The overall idea behind this article is to design the pressure sensor skin for low cost prosthetic hand, which is practical, environment friendly and scalable. Herein, the mathematical modelling, FEA, characterization and application of paper-based touch mode capacitive pressure sensor is presented. The sensor operates in normal, transition, touch and saturation region. The capacitance versus pressure plot shows linearity and sensitivity of the sensor. The sensor is operated and experimentally tested for all different modes: normal, transition, touch and saturation. The sensitivity and linearity of the fabricated capacitive pressure sensor is calculated and discussed. The sensor is experimentally tested for air pressure monitoring from 1 to 55 kPa and the capacitance is measured using Keithley Semiconductor Characterization System (Model – 4200 SCS). The linear rotation in position servo motor causes the precise movement in robotic arm which has been controlled using the linear range of pressure sensing element which can be utilized for manufacturing the low-cost prosthetic hands/skins.

## II. MATHEMATICAL MODELLING

The plate/diaphragm deflects due to uniform pressure application (P). The deflection in the circular plate is given by the partial differential equation [40]:

$$\left(\frac{\partial^2 W}{\partial r^2}\right)^2 + \frac{1}{r^2}\left(\frac{\partial W}{\partial r}\right)^2 + \frac{1}{r}\frac{\partial^2 W}{\partial r^2}\left[2\frac{\partial W}{\partial r} - \frac{h}{D}\frac{\partial \varphi}{\partial r}\right] = \frac{P}{D} \quad (1)$$

where, h, D, Φ and W are diaphragm thickness, flexural rigidity of diaphragm, airy stress and diaphragm deflection at radius r respectively. The flexural rigidity is a function of Young's modulus of elasticity, diaphragm thickness and Poisson's ratio of diaphragm material. If the circular diaphragm, made of elastic, homogeneous and isotropic material, is clamped at the edges, then after applying the boundary conditions, the deflection in the diaphragm at any distance r due to uniform pressure application is given by [10, 19]:

$$W(r) = W_0 \left[1 - \left(\frac{r}{R}\right)^2\right]^2 \quad (2)$$

where, W0 is the diaphragm deflection at the center and R is the diaphragm radius. A large deflection due to application of pressure in circular diaphragm is given by [19-20]:

$$W_{0,l} = \frac{PR^4}{64D}\left[\frac{1}{1 + 0.488\frac{w_{0,l}^2}{h} + \frac{\sigma h R^2}{16D}}\right] \quad (3)$$

where, h, σ, D, P are diaphragm thickness, build-in-stress, flexural rigidity and applied pressure respectively. A small deflection in circular diaphragm with build-in stress is given by [10]:

$$W_{0,s} = \frac{PR^4}{64D}\left[\frac{1}{1 + \frac{\sigma h R^2}{16D}}\right] \quad (4)$$

The base capacitance of the parallel plate capacitive pressure sensor is given by:

$$C_0 = \frac{\varepsilon_0 \varepsilon_r A}{d} \quad (5)$$

where, ε0 is permittivity of air, εr is permittivity of medium. The capacitance variation due to application of pressure on circular diaphragm (for both small and large deflection) is given by [10, 19-20]:

$$C_w = \int_0^{2\pi}\int_0^a \frac{\varepsilon_0 \varepsilon_r \, r \, dr \, d\theta}{d - W(r)} \quad (6)$$

The capacitance of touch mode capacitive pressure sensor due to large deflection in diaphragm is given by [from Eq. (2) and (6)]:

$$C_T = \frac{\pi\varepsilon_0\varepsilon_{t_1}a^2}{t_1} + \frac{2\pi\delta\varepsilon_{t_1}}{W_{0,l}}[b^2 A_1 + abA_2] \quad (7)$$

where,

$$\delta = \frac{\varepsilon_0 W_{0,l}}{t_1 + d\varepsilon_{t_1}} \quad (8)$$

$$A_1 = \ln\left|\frac{1+\sqrt{\delta}}{1-\sqrt{\delta}}\right| \quad (9)$$

and,

$$A_2 = \frac{1}{1-\delta} - \frac{2\delta}{3(1-\delta)} - \frac{\delta(3\delta+1)}{5(\delta-1)^3} \quad (10)$$

In MEMS, the composite membrane based circular shape capacitive and piezoresistive pressure sensor with silicon, silicon nitride and silicon oxide have been considered for mathematical modelling and simulations [41-42]. The MEMS based triple-layered square shape capacitive pressure sensor is presented with mathematical modelling, simulation and fabrication [43]. If the composite membrane is made of aluminum and polyimide [Al (E1, υ1, h1) and PI (E2, υ2, h2)], then the flexural rigidity of the diaphragm is given by [41-42]:

$$D = \frac{E_1[(h-e)^3 - (h_2-e)^3]}{3(1-v_1^2)} + \frac{E_2[(h_2-e)^3]}{3(1-v_2^2)} \quad (11)$$

where, e is neutral plane which is given by:

$$e = \frac{\frac{E_1}{1-v_1}h_1(h_1+2h_2) + \frac{E_2}{1-v_2}h_2^2}{2\left[\frac{E_1}{1-v_1}h_1 + \frac{E_2}{1-v_2}h_2\right]} \quad (12)$$

### III. FEM ANALYSIS

The thickness and available diaphragm radius for pressure sensing are 25 μm and 1 cm, respectively. The diaphragm radius is too large to perform finite element analysis. Hence, the diaphragm radius is scaled for validation using mathematical modelling. The FEM-simulation of 100 μm radius and 25.2 μm thick (200 nm Al and 25 μm PI) diaphragm is simulated for the pressure range of 0 kPa – 10 kPa. The comparison between mathematical modelling and simulation is presented [Fig 2(a)]. The von Mises stress and deflection in diaphragm is shown at 10 kPa [Fig 2(b) and (c)]. The deflection versus pressure variation is linear and follows the Hook's Law. The von Mises stress is maximum at clamped edges and deflection is maximum at the center of diaphragm. The CoventorWare® software is utilized to perform the FEM analysis. The meshing element type is the

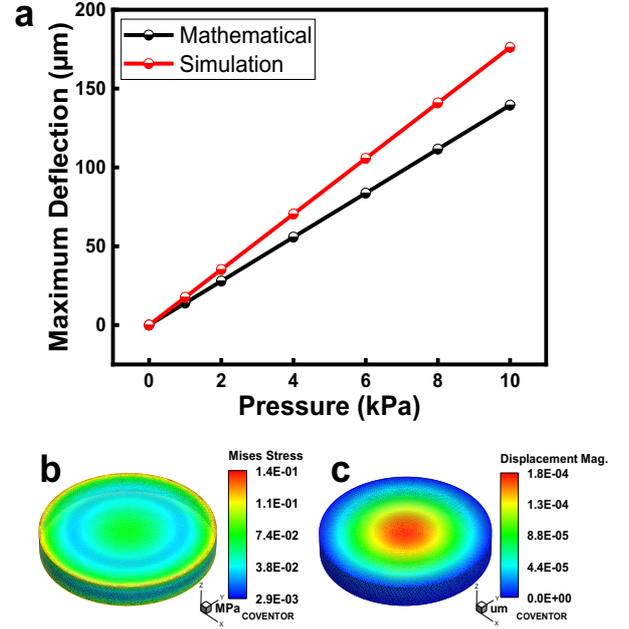

Fig. 2. (a) Validation of mathematical modelling and comparision with finite element analysis of circular dipahrgm deflection. The thickness and radius of diaphragm are 25.2 μm and 100 μm, respectively. (b) von Mises stress at 10 kPa (c). Diaphragm deflection at 10 kPa.

Tetrahedron parabolic with the element size of 6.3 using MemMech module.

### IV. FABRICATION PROCESS FLOW OF SENSOR

The fabrication process flow for the flexible capacitive pressure sensor is shown in Fig 3. The fabrication starts with cutting two pieces of circular shape from flexible aluminum (Al) sputtered polyimide (PI) sheets for bottom and top electrodes [Fig 3(a)]. These two pieces are of same radius which is 1.2 cm. First, a layer of double-sided tape is placed at the polyimide side of first piece which is back-plate of sensor [Fig 3(b)], followed by layering of double-sided tape two times more [Fig 3(c)]. The width of single layer double-sided tape is 2 mm and thickness is about 95 μm. After using three layers of double-sided tape, another piece of circular shape is placed at the top which will behave as pressure sensitive diaphragm, however, in this case the PI layer is at the top side of sensor [Fig 3(e)]. The digital image of the actual fabricated capacitive pressure sensor is shown in Fig 3(f), bending due to compressive stress. The three layers of double-sided tape are used to separate and clamp both the electrodes so that air can be used as a dielectric layer. The achieved separation gap is about 400 μm which increases from the actual separation gap after three layers of double-sided tape due to its adhesive nature from both sides. The Scanning Electron Microscopy (SEM) image of the sensor shows all the three layers of stacked tape [Fig 3(g)] and the inset is showing the pressure sensitive diaphragm [(Fig 3(f)]. The thickness of the diaphragm is 25 μm, separation gap is 412 μm at the edges and 390 μm at the center, which is due to the effect of build-in stress in diaphragm.

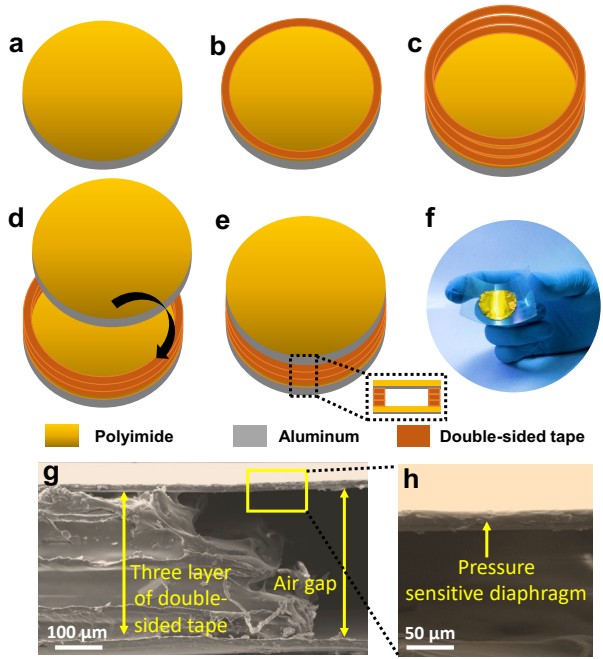

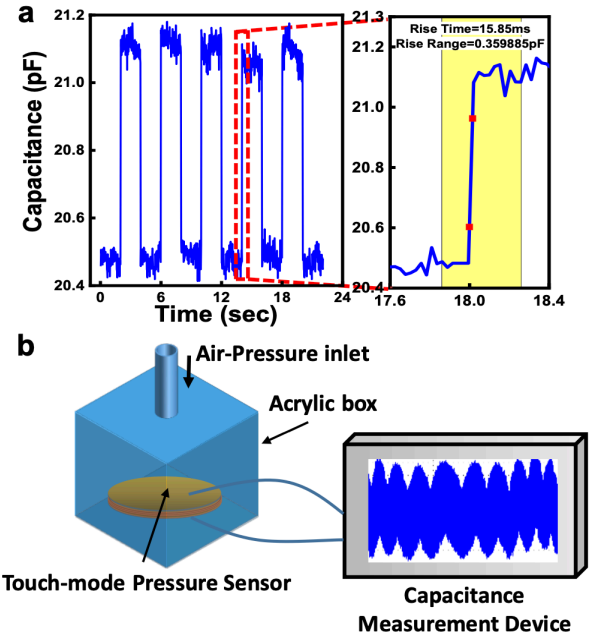

Fig. 3. Process flow for fabrication of flexible touch mode pressure sensors: (a) Cut the two circular pieces of from 25 μm thick Al coated PI sheet for top and bottom electrode. (b) Place a single layer of double-sided tape ring to clamp the edges. (c) Repeat the number of layers depending upon needed airgap. (d) Adhere top electrode to the third layer by double-sided tape. (e) Schematic of fabricated sensor. (f) Photographic image of fabricated capacitive pressure sensor on a Polyethelene terephthalate (PET) sheet to shown to admit the flexibility of sensor under the compressive stress. (g) SEM image of pressure sensor showing three layer of tape and air gap. (h) The zoomed part shows the diaphragm of sensor which is a pressure senitive mechanical element.

Fig. 4. (a) Repeated response of sensor for capacitance variation due to application of same pressure from 0 to 23 seconds. The sensor has rise time of about 15.85 ms and rige range of 0.36 pF. (b) Schematic diagram of experimental measurement setup for the cavacitance change due to deflection in the mechanically sensitive diaphragm due to application of pressure.

## V. EXPERIMENTAL ANALYSIS

The fabricated pressure sensor operation was tested by placing a 289 milligrams object at the center of the diaphragm repeatedly. The results of the pressure against time can be seen in Fig 4(a) which has the rise time of 15.85 ms. It can be seen that the weight of the object deflects the diaphragm to reduce the air gap between the parallel plates of the capacitive sensor. This, in turn, results in the change in capacitance. The consecutive peaks have almost the same capacitance change, which shows the reliability of the Al-PI foil/sheet as a material for diaphragms. In order to validate the mathematical analysis and FEM simulation, we carried out an experiment to find the response of the sensor in touch mode. The response time of the touch mode pressure sensor is 15.88 ms, which is less than the hollow-sphere micro-structure resistive pressure sensor [37], the graphene-based pressure sensor [38] and flexible-foam capacitive pressure sensor array [39]. A custom experimental setup was created to reach high pressures on the diaphragm to reach the touch mode. We placed the sensor on the bottom of an acrylic box. An air pressure nozzle was inserted on the top of the box such that the outlet of the nozzle is directly above the diaphragm. The experimental setup can be seen in Fig 4(b).

A custom scale from 0-100 was drawn on the knob of the nozzle to find out how much pressure the nozzle can apply with respect to each value on the scale. A commercial MEMS capacitive pressure sensor (MS5803-14BA) was used to calibrate the value of custom scale to the corresponding value of pressure. The commercial pressure sensor was placed under the opening of the nozzle where the sensor has to be placed. The knob was opened from 0-100 while noting down the pressure values from the commercial sensor. The results are shown in Fig 5(a).

It is evident that the knob of the air source results in a linear increase in pressure. Now, the fabricated sensor was placed under the nozzle and the knob was opened from 0-100 while taking the reading at intervals of 10. The results are shown in Fig 5(b). It can be observed that the response of the sensor starts as non-linear in the pressure range of (from 1-8 kPa). This is expected from a capacitive sensor in normal range of operation as long as the deflection in diaphragm is less than 1/3rd of the separation gap and follows the Kirchhoff's plate theory of deflection. As we reach large pressure values, the sensor goes into transition mode (8–10 kPa). In this mode, the pull-in phenomena occur and the diaphragm touches the back plate of the sensor. After that, further increase in pressure, transfers the sensor into touch mode operation in which response becomes linear for a wide range (from 10-40 kPa) and the capacitance increases as the large area of diaphragm touches the back plate. The response saturates thereafter as the maximum portion of the diaphragm is now stuck to the bottom plate.

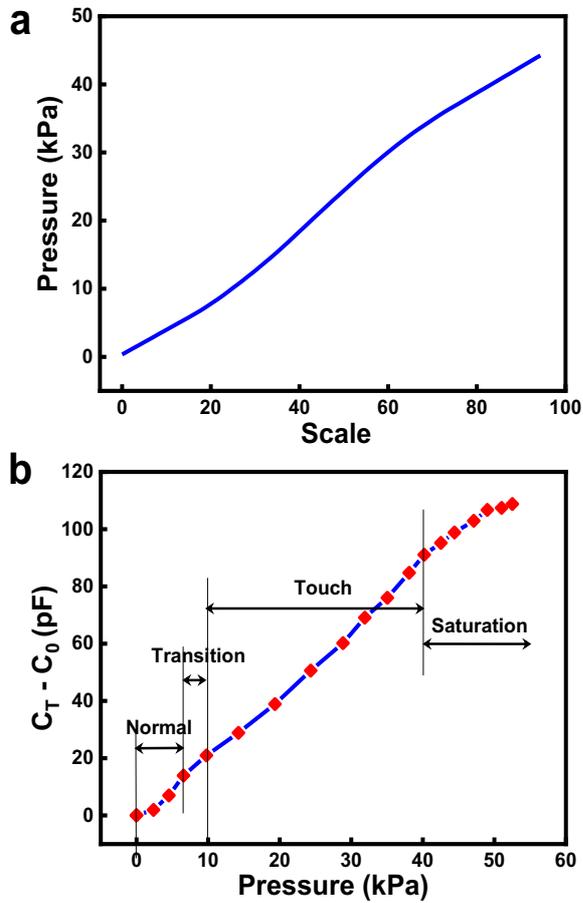

Fig. 5. (a) Response of commercially available MEMS capacitive pressure sensor used to calibrate the fabricated touch mode capacitive pressure sensor. (b). Response of paper-based touch mode capacitive pressure sensor for 0-60 kPa pressure range. The normal, transition, linear and saturation mode of flexible capacitive pressure sensor shows the linear response of our sensor.

## VI. Application

We present an application of our paper-based touch sensor in the form of a wireless servo motor control demonstration. Position based servo motors are generally used in robotic arms to undertake precise movements. The linear range and highly sensitive touch mode capacitive pressure sensor allows precise control of robotic arm movement for variety of applications. The paper-based touch mode pressure sensor is connected to a BLE PSoC that can read the capacitance from the sensor directly using its internal CapSense module as shown in Fig 6(a). This BLE PSoC then transmits data wirelessly to another BLE PSoC, which is controlling the position servo motor. The sensing end is compact and can be transformed into an electronic skin by adding more sensors to enhance spatial quantity of sensors. As the pressure is applied to the sensor in the linear range operation, the position of servo motor changes linearly. In Fig. 6(b), the paper-based touch mode sensor can be seen with the BLE chip visible from underneath PI substrate (35 μm thick) to the left of sensors. When pressure is applied on the sensor, this BLE chip sends the corresponding change in capacitance to another BLE chip that is connected to a servo motor as shown in the right side of Fig 6 (b). This setup opens up the possibility of a new way to control robotic arms. This pressure sensor can easily be converted into robotic/electronic skin like patch that can control a position-controlled servo that is used in robotic arms. As we apply pressure in which the response of sensor is linear, the servo motor rotates from $0^0$ to $90^0$ linearly [Fig 6(c)]. This range can be easily increased by scaling up the input to the servo motor in proportion to the change in capacitance. Fig. 6(c) shows the output plot of the change in capacitance that results in a linear change in rotation angle of servo motor from $0^0$ to $90^0$ in response to linear application of pressure in range of 10-40 kPa.

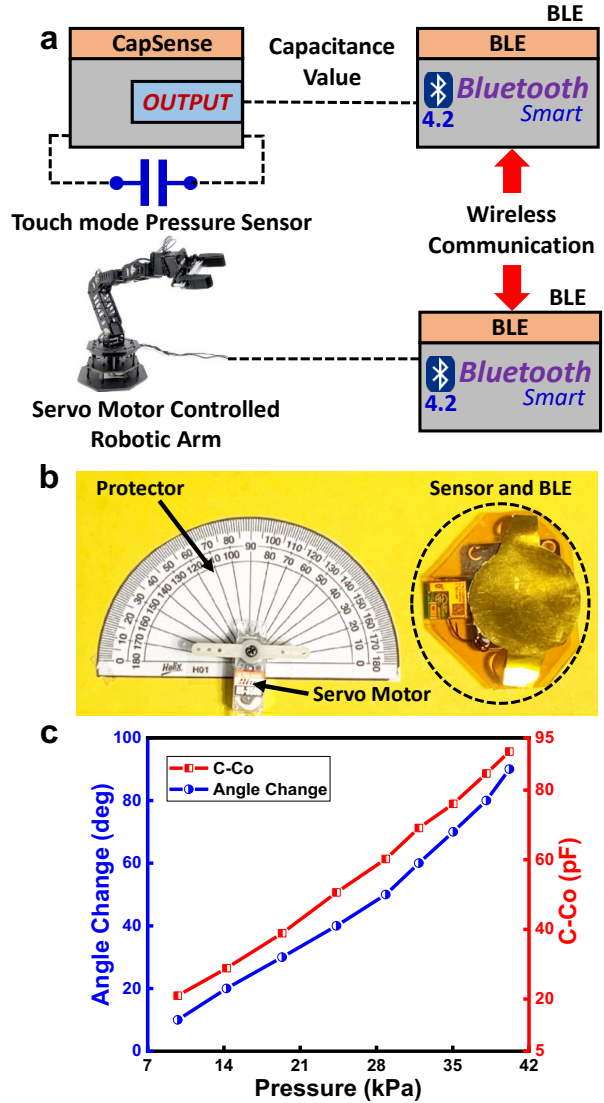

Fig. 6. (a) Schematics of pressure sensor for controlling the position of servo motor for precise movement in robotic arm using wireless signal communication. (b) Experimental setup for measuring the lienar change in angle variation due to linear pressure application. The protector is used for measuring the angle varation due to pressure application. The BLE chip is placed at underneath the Polyamide sheet with BLE chip for capacitance change transmition. (c) Plot for linear change in angle from $0^0$-$90^0$ and cpatitance change with pressure range from 10 kPa-40 kPa in which the response of sensor is linear. The trend in capacitance change and angle variation is maching with eatch other.

## VII. CONCLUSION

The mathematical, FEM and experimental analysis of paper-based touch mode capacitive gauge pressure sensor is presented. The mathematical modelling consists of deflection and capacitance variation with respect to pressure considering large deflection theory and built-in stress of pressure sensitive edge clamped diaphragm. The FEM analysis is presented after the geometric scaling of actual sensor for 0-10 kPa. The deflection in diaphragm is maximum at the sensor and von Mises stress is maximum at edges, for all five different steps after parametric variation of pressure. The diaphragm deflection follows the Hook's law of deflection theory which is linear in small deflection due to pressure application. The sensitivity and non-linearity of sensor is discussed after performing the capacitance versus pressure measuring experiments for air pressure sensing setup. the precise movement/displacement in a robotic arm, the linear range of pressure sensor is utilized for rotation in position of the servo motor as an application of sensor which is utilized for low-cost prosthetic hands. The sensor is fabricated using DIY- techniques which allows low-cost method for sensor fabrication for high range of pressure measurements for robotics, automation, aerospace, biomedical, consumer and portable electronics applications. For future work, the linear range of sensor will be increased so that more precise applications can be demonstrated.